\documentclass[letterpaper, 10 pt, conference]{ieeeconf}

\IEEEoverridecommandlockouts
\usepackage{cite}
\usepackage{amsmath}
\usepackage{graphicx}
\usepackage{algpseudocode}
\usepackage[]{algorithm2e}
\usepackage{textcomp}
\usepackage{amsmath}
\usepackage{pdfpages}
\usepackage{mathptmx}
\usepackage{enumerate}
\usepackage{url}
\usepackage{comment}
\usepackage{orcidlink}
\usepackage[utf8]{inputenc}
\graphicspath{ {figures/} }
\usepackage{caption}
\usepackage[skip=0cm,list=true,labelfont=it]{subcaption}

\usepackage{url}

\usepackage{breakurl}
\usepackage[vskip=1em,font=itshape,leftmargin=2em,rightmargin=2em]{quoting}

\usepackage[pagewise]{lineno}
\usepackage{graphicx}
\graphicspath{{./figures/}}

\usepackage{tabularx}

\title{\LARGE \bf
See Spot Guide: Accessible Interfaces for an Assistive Quadruped Robot
}

\author{Rayna Hata$^{1}$\orcidlink{⟨0000-0001-9682-1097 ⟩} , Narit Trikasemsak$^{2}$ \orcidlink{⟨0009-0006-0280-1305 ⟩}, Andrea Giudice$^{3}$, Stacy A. Doore$^{2}$ \orcidlink{⟨0000-0002-8322-6798 ⟩}
\thanks{$^{1}$ Rayna Hata is with the Robotics Institute, Carnegie Mellon University, Pittsburgh, PA, USA
        {\tt\small rhata@andrew.cmu.edu }}%
\thanks{$^{2}$Narit Trikasemsak and Stacy A. Doore are with the Department of Computer Science, Colby College, Waterville, ME, USA
        {\tt\small (Ntrika25,sadoore)@colby.edu}}%
}

\begin{document}

\maketitle
\thispagestyle{empty}
\pagestyle{empty}

\begin{abstract}

While there is no replacement for the learned expertise, devotion, and social benefits of a guide dog, there are cases in which a robot navigation assistant could be helpful for individuals with blindness or low vision (BLV). This study investigated the potential for an industrial agile robot to perform guided navigation tasks. We developed two interface prototypes that allowed for spatial information between a human-robot pair: a voice-based app and a flexible, responsive handle. The participants (n=21) completed simple navigation tasks and a post-study survey about the prototype functionality and their trust in the robot. All participants successfully completed the navigation tasks and demonstrated the interface prototypes were able to pass spatial information between the human and the robot. Future work will include expanding the voice-based app to allow the robot to communicate obstacles to the handler and adding haptic signals to the handle design.

\end{abstract}


\section{Introduction}

Guide dogs are one of the most trusted forms of assistive navigation for those in the blind-low vision (BLV) community. Approximately 20,000 guide dogs help BLV individuals successfully operate in the world by navigating unfamiliar and complex surroundings on a daily basis\cite{IGDF}. The benefits of guide dogs have been well documented with most studies observing that guide dogs provide additional social support for their handlers in the form of companionship, bonding, and protection\cite{Whitmarsh}. While working with a highly trained guide dog has many benefits, there are also significant responsibilities and limitations. This includes the high costs of training and care, travel restrictions, and a limited working span of about 6 years. Handlers must also keep applying for and training with new dogs as life circumstances and assistance needs change with age\cite{Clovernook}.

As assistive technologies have advanced in recent years, many researchers are experimenting with a way to address some of the challenges presented by guide dogs, not as a replacement but as an additional tool or resource for BLV individuals who may occasionally require an alternative or temporary navigation assistant. These types of devices include smart canes \cite{Chaitrali}, smartphone navigation applications using computer vision systems\cite{SeeingAI}, OSM \cite{OpenStreetMap}, GoodMaps\cite{GoodMaps}), and there also competitions demonstrating the performance of these systems for everyday tasks and navigation challenges\cite{Cybathlon}.

Robot navigation assistants could be especially useful in indoor settings such as airports, convention halls, and other public spaces where navigation can be more difficult than in outdoor settings. While there are several research teams working on different types of robotic guides, this paper reports on the results of a  study that investigates the potential for using industrial quadruped robots as navigation assistants. The present study focuses on the following research questions: 

\begin{enumerate}
  \item \textit{ What are the necessary features for a multisensory interface to facilitate critical human-robot interactions for blind navigation in a multilevel indoor environment?
}
  \item \textit{What human-dog pair navigation behaviors and tasks can be reproduced with an industrial quadruped robot?
}

\end{enumerate}

This paper investigates the potential for industrial quadruped robots to assist humans in executing similar navigation tasks that human-guide dog pairs perform every day. This study is grounded in a co-design framework\cite{Bird} and uses value-sensitive design (VSD) principles and methods\cite{Friedman} to design a multisensory interface prototype for the use of these robots in blind navigation tasks. We will first review the foundational skills and behaviors that must happen in successful human-guide dog pairs, the limitations of real-time human-dog communication in navigation tasks, and current research on using robots in blind navigation.Together with a co-author Andrea Giudice, a BLV Assistive Technology Specialist and guide dog handler, we co-designed and prototyped two interface prototypes to interact with a commercial industrial agile robot with the goal of developing an embodied navigation assistant. We report the preliminary results of a human-robot pair study with sighted participants and the agile robot operating as a team performing a variety of assistive navigation and retrieval tasks. 

\section{Related Works}

\subsection{ Blind and Low Vision Navigation }

For the 260 million individuals worldwide who have profound vision impairment, non-visual navigation can be a daily challenge as one must learn to rely on other sensory information to explore and navigate the world\cite{IABP}. Most individuals without the use of functional vision will go through Orientation and Mobility (O\&M) training to learn how to navigate independently and safely. This includes training on how to determine knowledge of intersection geometry, the state of the traffic signal based on traffic flow analysis, and awareness of nearby landmarks to understand surroundings. O\&M training helps individuals with BLV to use spatial reasoning and their navigation tools to navigate to their destination\cite{Giudice}. However, O\&M training is typically focused on learning a specific location, route,  or area, and often requires a BLV navigator to revisit O\&M training when moving to a new location. Common assistive approaches to non-visual navigation include using white canes, guide dogs, human assistants, environmental cues, or a combination of approaches depending on the navigation context\cite{APS}.

Modern use of guide dogs for non-visual navigation began in Germany during World War I, where the first guide dogs helped veterans blinded in combat. U.S. public knowledge of guide dog use has increased since the 1920s after a popular movement helped to establish the rights of handlers and dogs being allowed to operate freely in public spaces and the first guide dog training school in the U.S.(Eustis, 1927). There are approximately 20,000 official guide dogs in the U.S. with a greater demand than can be met by guide dog organizations. While guide dog schools in the U.S. provide their services free of charge to qualified BLV individuals, the cost of breeding, raising, health care, and ongoing training of the dog and their handler plus ongoing support ranges from \$40,000-60,000\cite{Clovernook}. 

While there is no standard size or breed for a guide dog, the dog must be appropriately sized and matched with its handler. Size is important because the height of the dog at the shoulder, plus the length of the harness, must fit comfortably with the height of the handler. The most common breeds used as guide dogs tend to be large in size such as German Shepherds (27–40 kg and 558–660 mm high) and Labrador/Golden Retrievers (27-36 kg and 558–635 mm high) because of their size and weight range.

\subsection{Quadruped Robots for Navigation Assistance}
Experimentation with robots as navigation assistants has increased over the last ten years as robots have become more agile and easier to operate independently. Many of the current research programs investigating the use of quadrupedal robots for assistive navigation utilize small to midsize robots to simulate interactions between handler-dog pairs. In \cite{Hwang} the researchers use a midsize quadruped robot, a Unitree Go1 model \cite{UnitreeGo1}, which is equipped with a custom handle that is able to receive directional commands from the user. Using the commands, the robot then plans a path around the obstacles to get to the next point and guides to user. Preliminary tests showed that the system is functional in most settings. Another study \cite{Kim}, also uses a similar mid-sized quadruped robot, a Unitree Aliengo\cite{Aliengo}, as a navigation tool designed for assisting BLV individuals.  This study developed a delayed harness model that tries to mimic similar interactions that transpire between a human and their guide dog. Reported results suggested accurate behavior prediction and better navigation performance with this handle design when testing blindfolded sighted participants. 

\cite{Xiao} implements the use of the Mini-cheetah robot\cite{Minicheetah} to create a robotics guide dog navigation system paired with a flexible leash system. The purpose of the leash is to get rid of the constraints that a rigid leash would bring in tight corners or narrow hallways. The leash is able to switch between a taut position and a slack position depending on the situation. They were able to successfully complete their study on blindfolded users, having them navigate through cluttered spaces. 

While having a slack leash may help improve movement in tight corners or spaces, the lack of rigidness diminishes the effectiveness of the communication between the user and the guide dog. Furthermore, in all of the studies above,  the size of the robots is a good deal smaller than the most common guide dog breeds (Labradors, Retrievers, Shepherds). Smaller to mid-sized agile robots used in these non-visual navigation studies require a long handle to reach between the human-robot resulting in awkward positioning when compared to a typical human-guide dog team, limiting body communication based on traditional training protocols. 

\subsection{Boston Dynamics Spot Explorer }

The Boston Dynamics Spot Explorer robot is an industrial quadrupedal, agile robot developed in 2016, and made commercially available in 2019. The Spot Explorer robot is considered large among the commercially available quadrupeds \cite{BDSpot}. This means the Spot model is approximately four times the size of the MIT mini Cheetah and two times the size of the Unitree robots used in the studies noted above. The Spot Explorer robot houses five stereo cameras and LIDAR. The standard Spot Explorer robot can move at a quick pace with advanced mobility and perception features to navigate stairs and rough terrain while collecting data in real time to avoid obstacles. The robot can be customized to move beyond its five standard data processing abilities of perception, computation, autonomy, integration, and manipulation. (See Table \ref{tab:robot-specs} for a comparison of robots).

\begin{table*}
\centering
\caption{Specifications of Small, Medium, and Large Agile Robots for Guided Navigation}
\begin{tabularx}{\linewidth}{|c|X|X|X|X|X|X|X|X|}
\hline
\textbf{Robot} & \textbf{Length (mm)} & \textbf{Width (mm)} & \textbf{Height (mm)} & \textbf{Weight (kg)} & \textbf{Speed (km/h)} & \textbf{Load (kg)} & \textbf{Degrees of Freedom} & \textbf{Battery Run Time (minutes)} \\
\hline
Mini-Cheetah~\cite{Minicheetah} & 480 & 270 & 300 & 9 & 8.8 & 10 & 12 & 120 \\
\hline
Unitree Go1~\cite{UnitreeGo1} & 588 & 220 & 290 & 12 & 12 & 5 & 12 & 120 \\
\hline
Unitree Aliengo~\cite{Aliengo} & 650 & 310 & 600 & 21 & 5.4 & 10 & 12 & 240 \\
\hline
Spot Explorer~\cite{BDSpot} & 1100 & 500 & 660 & 32 & 5.8 & 31 & 12 & 90 \\
\hline
\end{tabularx}
\label{tab:robot-specs}
\end{table*}

\begin{figure*}[htbp]
  \begin{minipage}{\linewidth}
    \centering
    \begin{minipage}{0.25\linewidth}
        \centering
        \resizebox{!}{0.75in}{\includegraphics{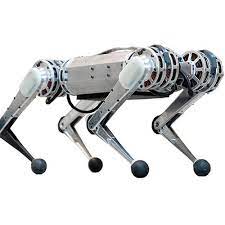}}
        \subcaption{Mini-Cheetah}
    \end{minipage}%
    \begin{minipage}{0.25\linewidth}
      \centering
      \resizebox{!}{0.85in}{\includegraphics{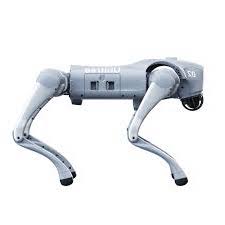}}
      \subcaption{Unitree Go1}
    \end{minipage}%
    \begin{minipage}{0.25\linewidth}
      \centering
      \resizebox{!}{1in}{\includegraphics{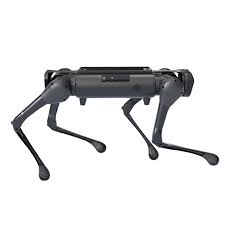}}
      \subcaption{Aliengo}
    \end{minipage}%
    \begin{minipage}{0.25\linewidth}
      \centering
      \includegraphics[width=\linewidth]{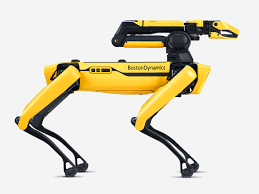}
      \subcaption{Boston Dynamics Spot}
    \end{minipage}
  \end{minipage}
  \caption{Commercially available quadruped robots used in nonvisual navigation research from smallest to largest: a) Mini-cheetah, b) Unitree Go 1 and Go 2, c) Unitree Alien Go, and d) Boston Dynamics Spot Explorer (with arm).} 
\end{figure*}

The Spot Explorer models are most frequently used in industrial settings performing tasks in monitoring gauges, inspecting equipment for failure points, constructing maps of large factory complexes, and detecting harmful emissions in chemical and energy plants\cite{Reuters}.  Other uses for the Spot Explorer model that are less common but have a higher public profile have been used in advertising campaigns\cite{Pistachios}, fashion runways\cite{Fashion}, halftime shows\cite{Halftime}, and in some cases, police surveillance and patrol\cite{Police}. To our knowledge, the use of a Boston Dynamic Spot Explorer model as a navigation assistant is unique to this study. 

\section{Assistive Robot Guide Co-design }
The term co-design refers to the method of designing an assistive technology with individuals who bring their own lived experience to the design, development, and testing cycles \cite{Holmes}. The principle of co-designing is closely tied to the disability rights movement philosophy of “nothing without us”. Too often, assistive technologies are designed independently of the intended user community, who are asked too late (if at all) to test an engineered solution created by a team without any first-hand knowledge of lived experience challenges and concerns that can directly impact individual safety and well-being. In order to fully incorporate BLV lived experience and handler expertise, our team began the project by interviewing and then working through a co-design process with our co-author, a BLV guide dog handler who has trained and worked with seven guide dogs on a daily basis for over 30 years.

We have also incorporated a set of underlying responsible computing principles, prospective study methodologies, and design practices known as value-sensitive design (VSD)\cite{Friedman}. VSD requires that prospective design projects include working through a formal process to identify and communicate with direct and indirect stakeholders and their potentially conflicting values in the prospective design process. In this project, we identify BLV guide dog handlers as the direct stakeholders and the general public as indirect stakeholders who must navigate in the same public spaces as the robot navigation assistant (e.g., retail areas, airports, sidewalks, etc.). 

\subsection{Co-design Data Collection}
At the beginning of this process,  our team spent hours together observing human-dog interactions, discussing communication cues and signals, body-movement patterns, and targeted behavior rewards that our co-designer uses to reinforce good guiding behaviors with her young guide dog. Guide dogs must learn many skills to help their handler navigate complex indoor and outdoor environments such as walking at a constant pace and gait, walking backward if needed, avoiding obstacles, opening doors, navigating stairs, and cutting through crowds. Through these observations and interviews during the co-design process, the team discussed core guide dog behaviors in order to compare standard guide dog behaviors to the robot’s embedded functions and features. Comparing the basic navigation skills required of a guide dog with the Boston Dynamics robot’s existing features provides a set of simple measures to test the robot’s potential to serve as a non-visual navigation assistant (Table \ref{tab:RobotVDog}). Guide dogs also learn to ignore distractions that interfere with their duties and are trained to ignore compelling objects such as stray balls, food, other animals, and people who want to pet and talk with them while they are working. For a robot, these types of distractions do not present a problem.

Finally, while the focus of this paper may be on the viability of BLV human-robot pairings for navigation, we are in parallel considering the values, concerns, and safety of the indirect stakeholders (sighted public) in these spaces. Given the early stage of the interface designs, we wanted to observe sighted participants in a human-robot pair guiding and voice command tasks to assess their comfort around the robot and their ability to work with the robot to complete the protocol. Similar research is being conducted on reported levels of trust and safety using the same Boston Dynamics Spot Explorer model in experiments with sighted participants who are not guiding the robot but instead are directed to walk past the robot in a close contact setting\cite{UTPaper}. This study found there was a higher level of trust when the robot was wearing a designated service animal vest and when a human operator was walking with the robot.

\section{ Human-Robot Interaction Prototypes}
Our goal was to use the observations and input from our co-design partner in a preliminary study to assess the industrial robot’s potential as a navigation assistant and to compare the results to findings from previous studies that employed smaller robots with the same types of navigation tasks. Our first challenge was to design two prototype human-robot interaction interfaces that would allow the human-robot pair to communicate with one another during a set of simple navigation challenges: 1) a functional handle attached to the robot that would communicate body movements between the robot and the handler; and 2) a voice-based interface that would allow the handler to give voice commands to the robot through simple directions.  

\subsection{Prototype Handle Design}

\begin{figure}[!h]
    \centering
    \includegraphics[width=0.3\textwidth]{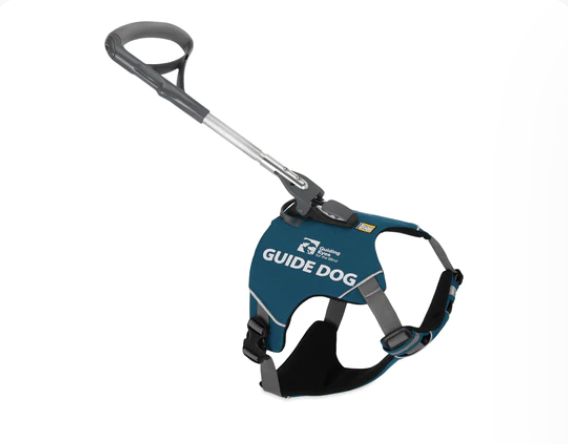}
    \caption{Ruffwear Guide Unifly dog harness}
    \label{fig:Ruffwear}
\end{figure}

\begin{figure}[!h]
    \centering
    \includegraphics[width=0.3\textwidth]{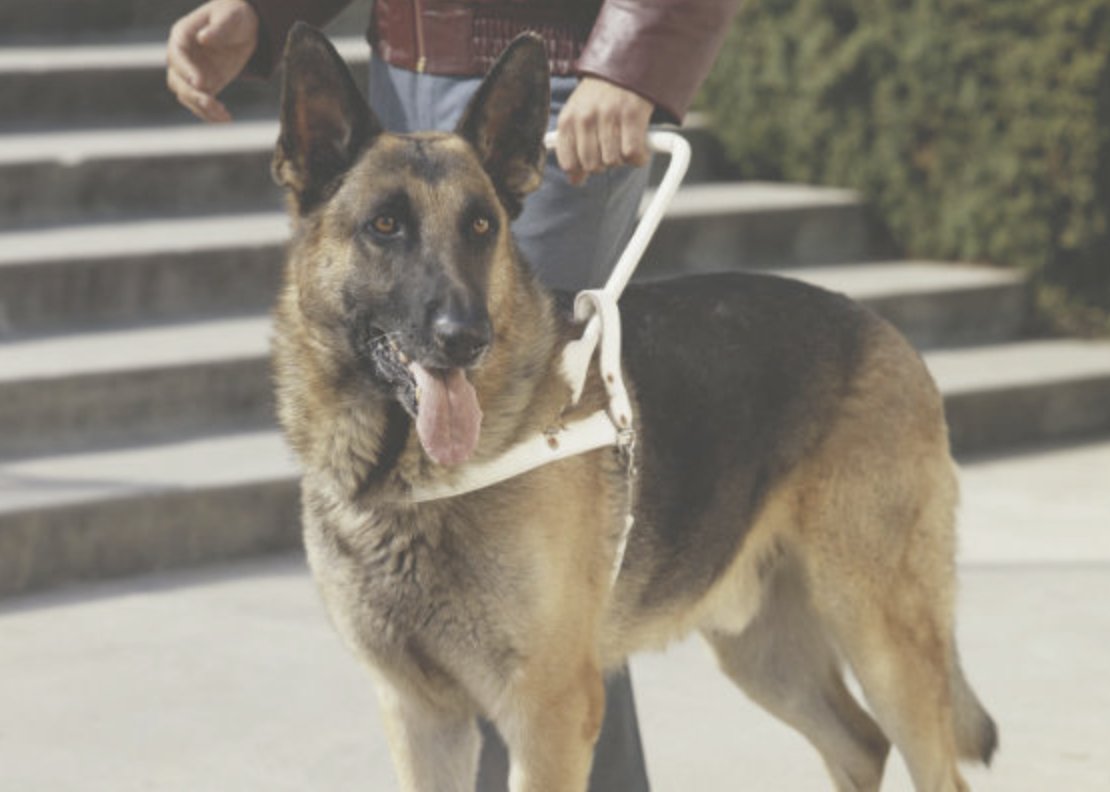}
    \caption{Standard Guide Dog Harness}
    \label{fig:GuideDogHarness}
\end{figure}

After observing the body placement of our co-design partner next to her guide dog, we recognized that the Spot Explorer dimensions and camera placements would require designing and building a modified handle to be placed on the rear half of the robot’s outer shell. The integrated cameras on the robot are required for object detection and avoidance during navigation. If the robot detects an obstacle (i.e., a human handler) blocking one of the cameras the robot will at first move away in order to avoid a collision. This would be counterproductive and dangerous if the robot was acting as a guide. The handle needed to be placed far enough back on the robot to avoid the right rear side camera but still near enough to the back center to provide clear body signals for the handler to understand and react in parallel to the robot’s movements. The design of the prototype handle was modeled off a newer style of harness\cite{Ruffwear} that is designed for BLV runners using their guide dogs (Figure \ref{fig:Ruffwear}). While not a typical guide harness system, it is becoming increasingly popular among guide dog handlers because of the greater freedom of movement and reactions it allows for the handler’s wrist. 

\begin{figure}[!h]
  \begin{minipage}{\linewidth}
    
    \begin{minipage}{0.3\linewidth}
      \centering
      \resizebox{!}{1in}{\includegraphics{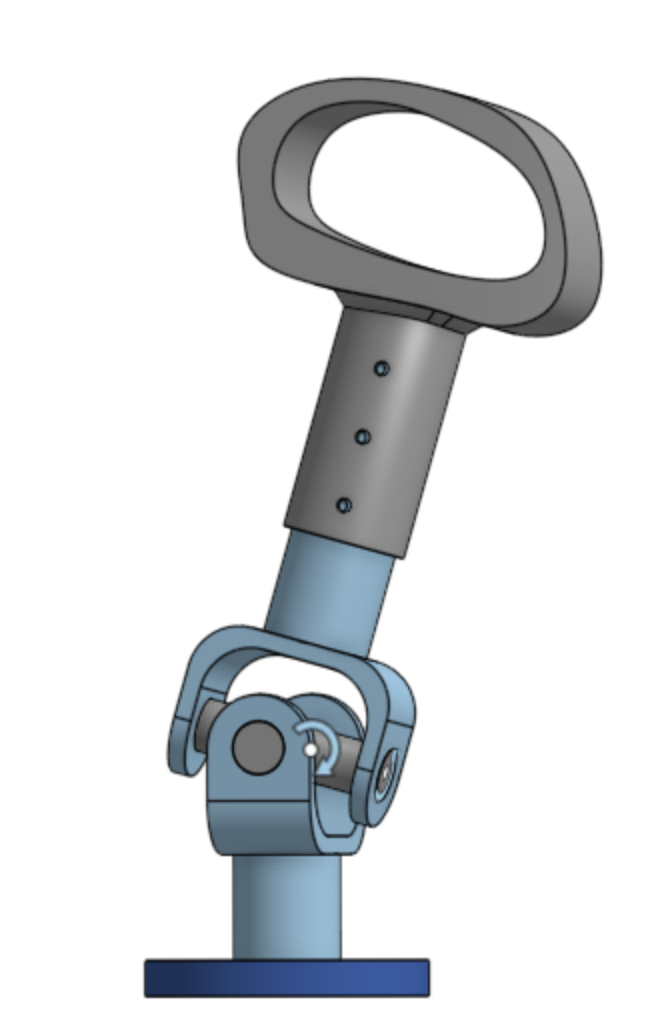}}
      \caption{Version 1}
      \label{fig:version 1}
    \end{minipage}%
    \begin{minipage}{0.3\linewidth}
    
      \resizebox{!}{1in}{\includegraphics{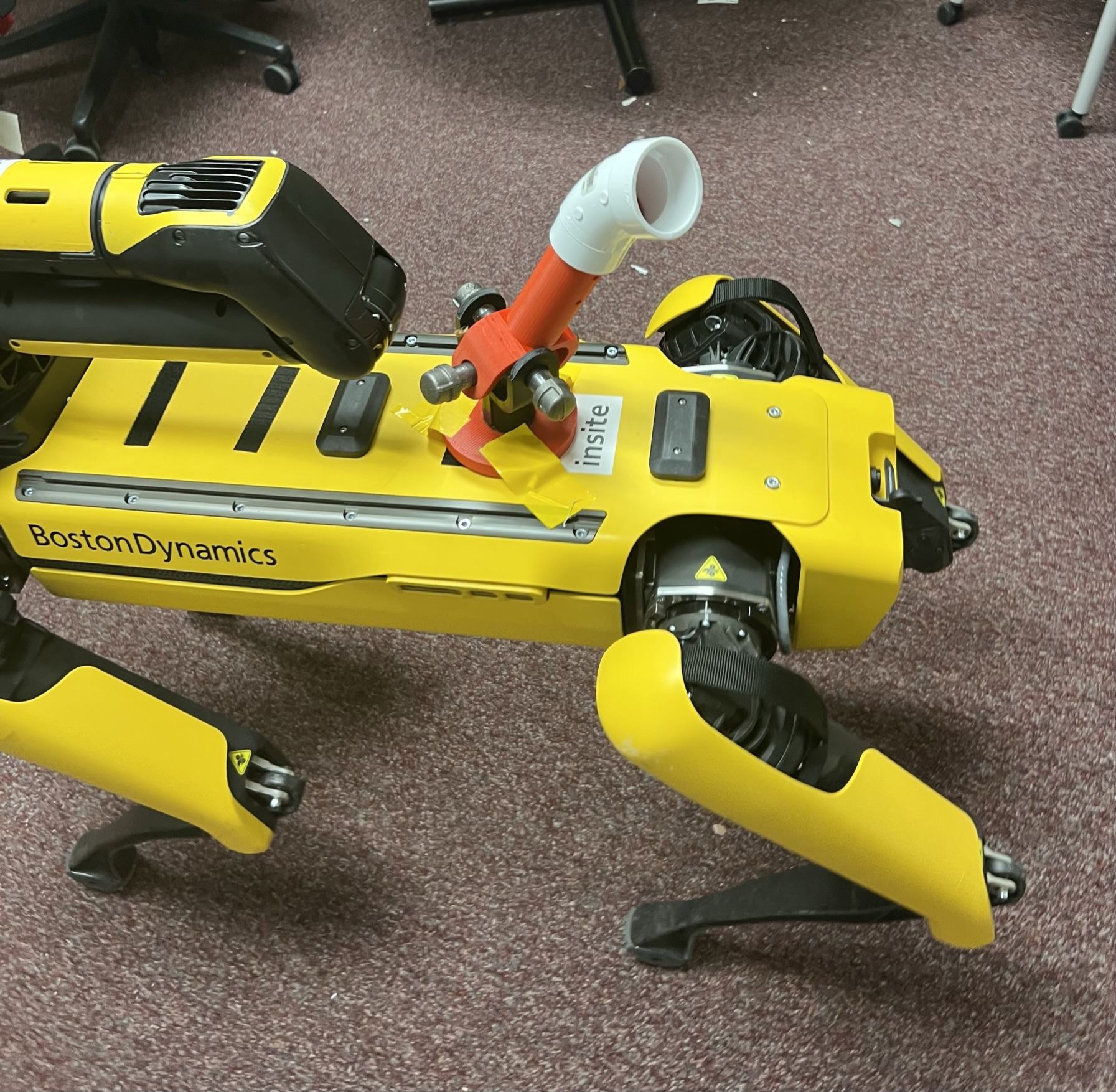}}
      \caption{Version 2}
       \label{fig:version 2}
    \end{minipage}%
    \begin{minipage}{0.3\linewidth}
      \centering
      \resizebox{!}{1in}{\includegraphics{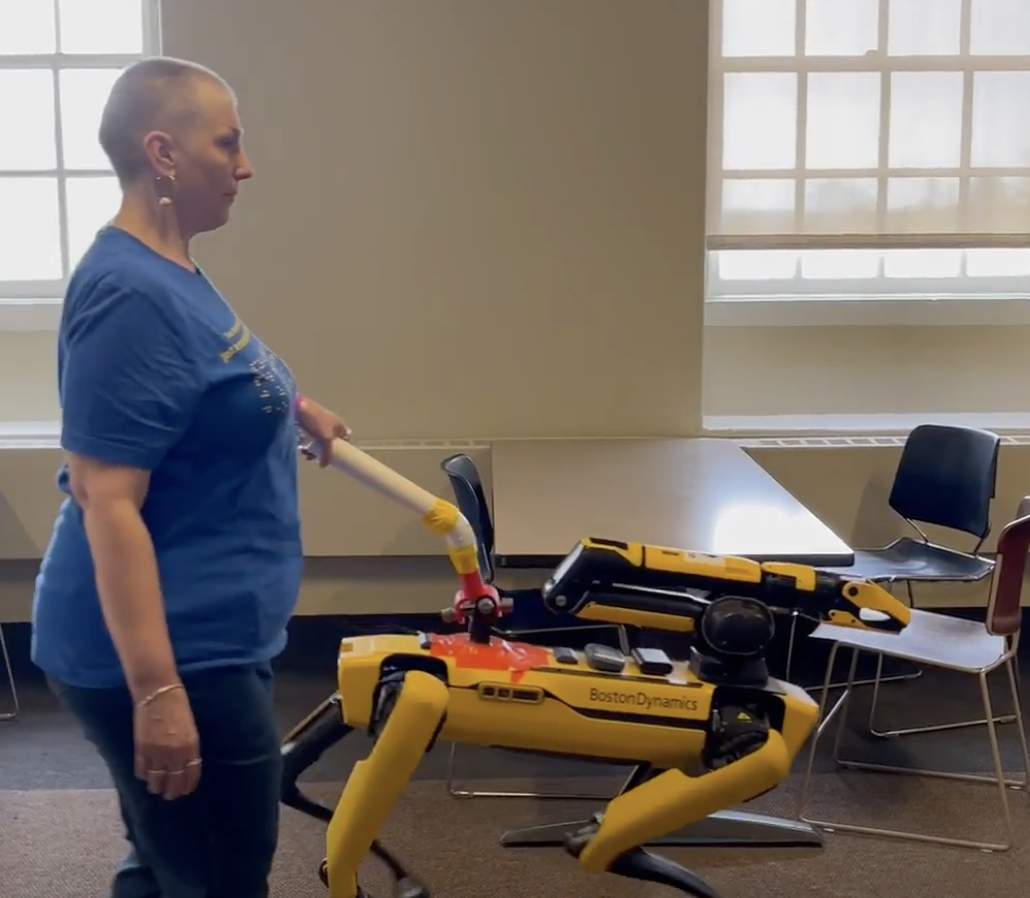}}
      \caption{Version 3}
       \label{fig:version 3}
    \end{minipage}%
  \end{minipage}
\end{figure}

Standard guide harnesses(Figure \ref{fig:GuideDogHarness}), as well as the Ruffwear harness, require a strap that would typically be placed around the guide dog’s chest and abdomen. The robot’s rectangular body structure, size, and cameras would not work with these harness systems because the straps would block the front and side cameras. Thus, we designed a custom handle without body strapping that could be positioned on the back of the robot. The first version of the guide handle was created using OnShape CAD software and was created using a 3D printer. The original design of the handle (Figure \ref{fig:version 1} consisted of two forks connected by a universal joint (U-joint) and a handle that was attached to the upper fork. The universal joint created a similar movement as the Ruffwear’s handle, allowing for a 120-degree turn on both vertical and horizontal movements. However, after some early testing, we found the 3D printed U-joint was not strong enough to withstand the continuous or sudden movement interactions between a human and the robot, snapping after a few rotations. 

Version 2.0 of the handle (Figure \ref{fig:version 2}) shared the same basic shape as version 1.0, however, we replaced the 3D-printed u-joint with one made from two metal bolts. The metal bolts proved to be more durable than the 3D-printed pieces and allowed for greater freedom of movement of the handler’s wrist. The printed shaft of the handle was replaced with a short PVC pipe with an elbow curve at the top. The handle was mounted at the rear to avoid the robot arm and to help the handler stay out of the right-side rear camera view.  During the next round of pilot tests, the handle proved to be too short causing the handler to drift into the camera’s view while turning. This created a safety hazard causing the robot’s actions to become increasingly erratic, putting the handler in harm's way. In order to rectify this issue, version 3.0 of the handle was attached to a longer shaft of PVC pipe (Figure \ref{fig:version 3}). With the additional length, the handler could remain outside the camera’s view yet still have their body close enough to the robot to understand and anticipate its guiding movements. The final user studies were conducted using the third iteration of the handle design. 

\subsection{ Prototype Voice-based Interface Design}

\begin{figure}[h]
    \centering 
    \includegraphics[width=0.45\textwidth]{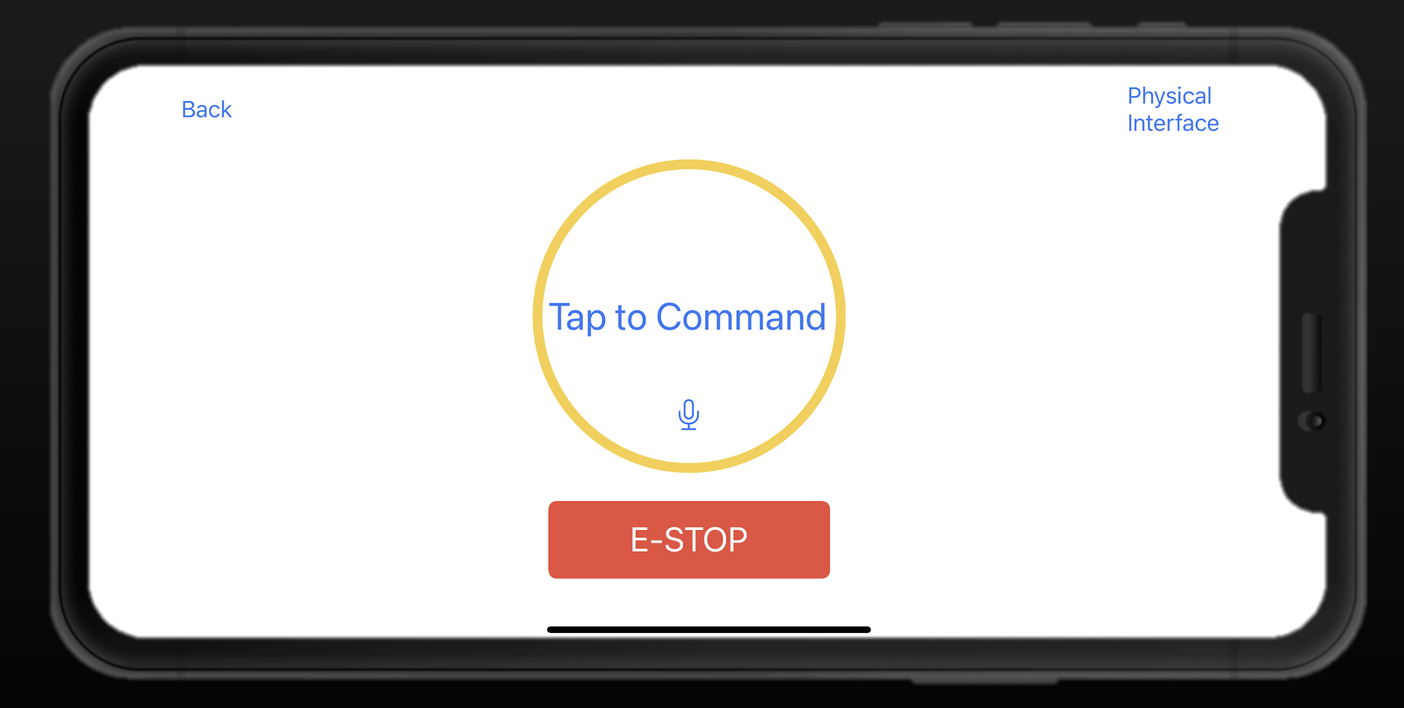} 
    \caption{Voice-based and Touchscreen Interface}
    \label{fig:Screen} 
\end{figure}

Guide dogs and their handlers communicate in a variety of ways. Some of this communication happens nonverbally with shifting of body positions, hesitations, longer pauses, and directional movements. In other cases, the handler gives direct verbal commands such as forward, or orientation (right, left), or follow-up commands such as “find railing” when the guide dog stops suddenly to indicate they are approaching a set of stairs. In order to provide a mechanism for the handler to use voice cues for communication, our team designed a prototype voice-based smartphone app that would allow the handler to give the robot a set of basic direction and orientation commands. The app was designed using Swift and UIKit to interpret the voice command and pass it to the robot’s network, which would in turn pass the pre-programmed functions to the robot (Figure \ref{fig:Screen}).

\begin{figure}[h]
    \centering
    \includegraphics[width=0.5\textwidth]{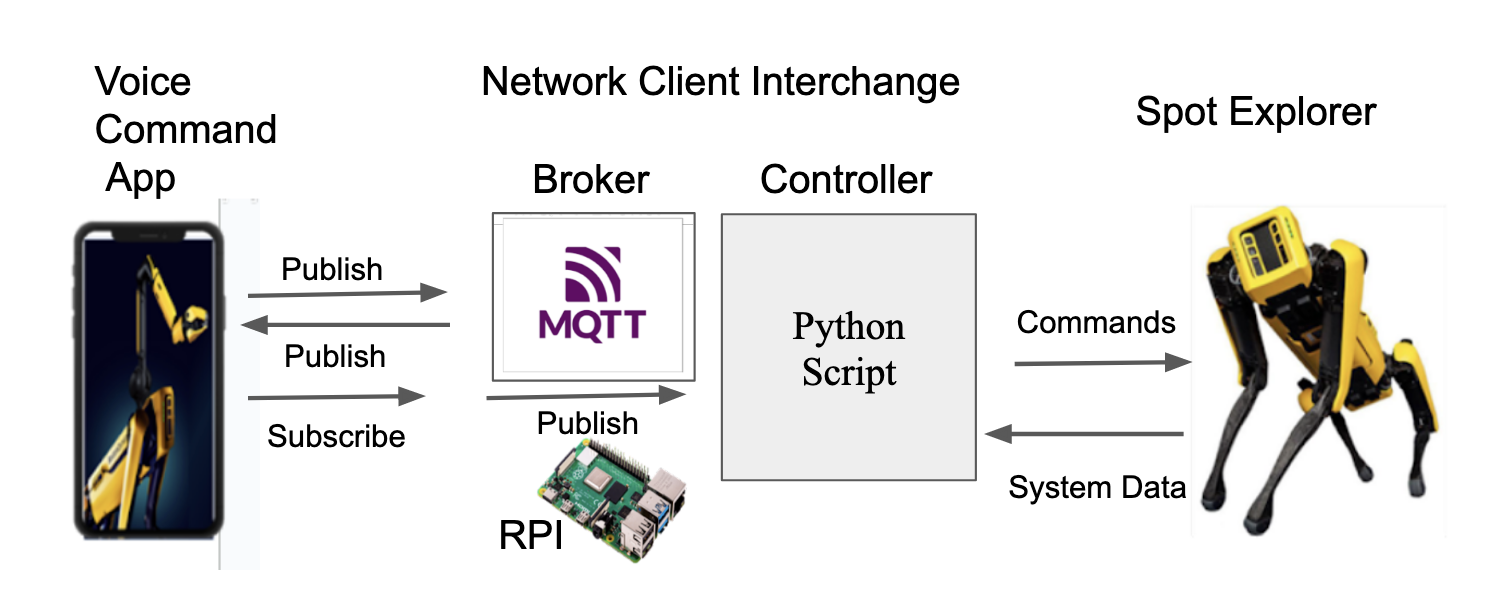}
    \caption{Voice Interface Network Architecture}
    \label{fig:RaspPI}

\end{figure}

The voice-based control prototype was designed with a Handler UI and a backend networked controller.  The frontend iOS app provided the handler with an accessible interface to pass simple verbal commands to the robot.  The backend served as the control system for Spot and managed all interactions with the robot.  In this preliminary study, we used the Raspberry Pi (RPI) to serve as the backend host and an iPhone 12 to take advantage of Apple’s built-in accessibility features. The two systems communicate with each other through Mosquitto (MQTT), a framework developed for IOT applications that has more recently been applied to robotics.  The RPI served as a MQTT broker as it managed all incoming messages and routed them, while the backend files acted as a client and subscribed to the broker, even though they were on the same device.  The iPhone also subscribed as a client, meaning both components could send and receive information (Figure \ref{fig:RaspPI}).

The prototype voice-based interactions were designed to use a combination of finger taps for initiating and sending the commands and voice to give the commands. The handler spoke the command slowly and clearly, using a short pause between words to help the app transcribe the commands correctly. The Apple Speech framework then transcribed the handler's command and displayed it on the screen in real time. The interface also had the option of using a start/send command button to accommodate individuals with residual vision who might want to control the text size and high-contrast features. If it was a valid command, the robot performed the action and stopped. For the first iteration of the prototype, the handler could give the following valid commands: 1) “ Spot, go forward” three steps forward and stops, 2) “Spot go backward”: three steps and stops, 3) “Spot go right” or “Spot go left”: three steps to 90 degrees from its original position.

\section{Study Design and Methods}
Our goal was to test both the handle and mobile app prototypes with a group of sighted participants and our guide dog expert/co-designer to assess: 1) how effectively the human-robot pair could work to complete a few simple navigation tasks, 2) how effectively each of the prototype communication interfaces (i.e., handle and app) worked independently, and 3) the handler's reported level of comfort and confidence in the robot’s interactions during the navigation tasks.

\subsection{Participants}
The user study consisted of a convenience sample of 20 sighted undergraduate students and faculty recruited from a liberal arts college through email invitations to the campus community. Participant demographics were not considered an important factor in the analysis, however, slightly more than half of the participants self-reported as female-identifying and the reported age of the sighted participants ranged from 18-60 years old. Participants in the sighted user study were able to walk independently around campus. While there was a range of reported previous exposure to the Spot Explorer robot, about half of the sighted participants reported brief encounters with the agile robot around campus, none of the participants had used the prototype handle or app or been guided by the robot prior to their participation in the study. A second single-participant study was conducted with our co-designer, a BLV guide dog handler, for 36 years. She completed the same practice procedure and navigation study protocol as the sighted participants. She also completed an additional outdoor navigation protocol designed specifically for her at her request. All participants completed a post-study feedback survey. Participants were study volunteers and were not compensated for their participation in the study. The user study protocol and post-study survey were reviewed and approved by the IRB and all participants indicated their formal informed consent prior to the start of the study.

\subsection{Protocol and Data Collection}
The user study protocol was divided into two parts. The first part consisted of a practice session to acclimate the participant to the robot’s movements while using the handle and the second part consisted of an autonomous walk through an indoor environment (i.e., library) where the participant was being guided by the robot through the use of the prototype handle. The practice session was designed to teach three separate human-robot pair movements: walking forward, turning left and right, and ascending and descending a short flight of stairs. The participants were told they could repeat any of the actions until they felt comfortable with executing the movements with the robot, although most participants did not request additional practice time. The researcher controlled each of the movements during the practice sessions and the participants were verbally notified of what action the robot would take before each part of the session. 

The navigation tasks were considered completed if a participant was able to complete the following tasks independently after the brief practice session: 1) the participant could successfully give four commands to the robot via the voice interface app, 2) the participant could use the handle to let the robot navigate 100-foot curved auto walk route and back to the origin location, and 3) the human-robot pair could navigate 20 feet to a short flight of stairs, ascend the four steps, follow the robot as it turned around descend the steps (i.e., safest way for this type of agile robot to descend) allowing the handle to pivot 180 degrees in their hand. Participants were also asked to complete a post-study survey about the handle and voice command app prototype's functionality and their levels of trust in the robot interactions during the trials. Our co-designer completed the same practice session and study protocol after completing most initial testing sessions with sighted participants. This was for safety reasons as well as scheduling accommodations.

The navigation tasks were performed using the robot's auto-walk feature, where the robot autonomously navigated a pre-planned route with the user. While the route was pre-planned, the robot needed to make real-time decisions based on new information such as obstacles or approaching humans in a busy library setting. The route contained the type of simple indoor paths that a handler-guide dog pair might need to navigate daily. To complete the multilevel navigation, the robot needed to lead the participant to the start of a short flight of stairs (4 steps to a landing area) while maintaining a straight path and a constant pace.

Approaching the stairs, the robot would stop and signal to the participant that they were about to go upstairs by lowering its rear down and tipping its front up so that the participant’s arm would get raised by the attached handle. After a 3-second pause, the robot would ascend the stairs and then pivot so that its back would be facing down the stairs that it had just climbed. Before descending down the stairs, the robot would once again pause and signal to the user that they are about to go downstairs by raising its back end and lowering its front end so that the participant’s arm would be pointed down. After the pause and signaling, the robot resumed moving to descend the stairs. After reaching the bottom of the stairs, the robot turned to the right to lead the participant along the rest of the hallway route and then turned around to move back to the starting point.

\subsection{Post Study Survey}
After completing the navigation tasks, participants were asked to fill out a survey ranking their level of agreement with statements about their experience. The three questions asked about the handle and the user’s comfort, another set of four questions asked about the functioning of the voice-based app, and the last 3 questions asked about whether the user felt comfortable being around the robot. The Likert scale was anchored from one to five, with 1= strongly disagreeing with the statement and 5=strongly agreeing with the statement. As this study is exploratory and descriptive in nature, the survey was designed to be short and only focused on the key prototype features and participant perceptions related to overall trust and safety interacting with the robot. 

\section{Results}
The user study navigation protocol results were measured by the successful completion of the navigation protocol after a brief training session. As this was a preliminary study of the prototypes, there were no measures of accuracy or time on task, instead, we measured success as if the participants (BLV Co-designer = 1, Sighted Participants = 20) could complete each navigation task using the handle prototype and if the participants could complete the voice command tasks using the voice-based interface (BLV Co-designer = 1, Sighted Participants = 20).

\subsection{Co-designer Responses and Feedback}
Our team spent hours together observing our co-designer demonstrating human-dog interactions, discussing communication cues and signals, body-movement patterns, and targeted behavior rewards used to reinforce good guiding behaviors with a young guide dog. Guide dogs must learn many skills to help their handler navigate complex indoor and outdoor environments such as walking at a constant pace and gait, walking backward if needed, avoiding obstacles, opening doors, navigating stairs, and cutting through crowds. Through these observations and interviews during the co-design process, the team identified core guide dog behaviors in order to compare standard guide dog behaviors to the robot’s embedded functions and features. 

Comparing the basic navigation skills required of a guide dog with the Boston Dynamics robot’s existing features provides a set of simple measures to test the robot’s potential to serve as a non-visual navigation assistant (Table \ref{tab:RobotVDog}). Guide dogs also learn to ignore distractions that interfere with their duties and are trained to ignore compelling objects such as stray balls, food, other animals, and people who want to pet and talk with them while they are working. For a robot, these types of distractions do not present a problem.

During the co-design process, we also learned about the potential limitations of the robot that would need to be adjusted or re-engineered to meet a basic level of navigation assistant functionality.  For example, guide dogs need to recognize curbs and stairs and avoid hazards from above and below, such as low-hanging limbs, power lines, and potholes. In addition to listening to their handler’s commands, guide dogs are trained to reason beyond the commands to perform acts of intelligent disobedience. This is when the dog understands something about the environment that the handler does not (e.g., a speeding car or sudden drop) and this is a critical safety skill for both the handler and the guide dog. This type of intuition training in unexpected situations would be difficult to train in a robot.

\begin{figure}[h]
    \centering 
    \resizebox{!}{2.5in}{\includegraphics{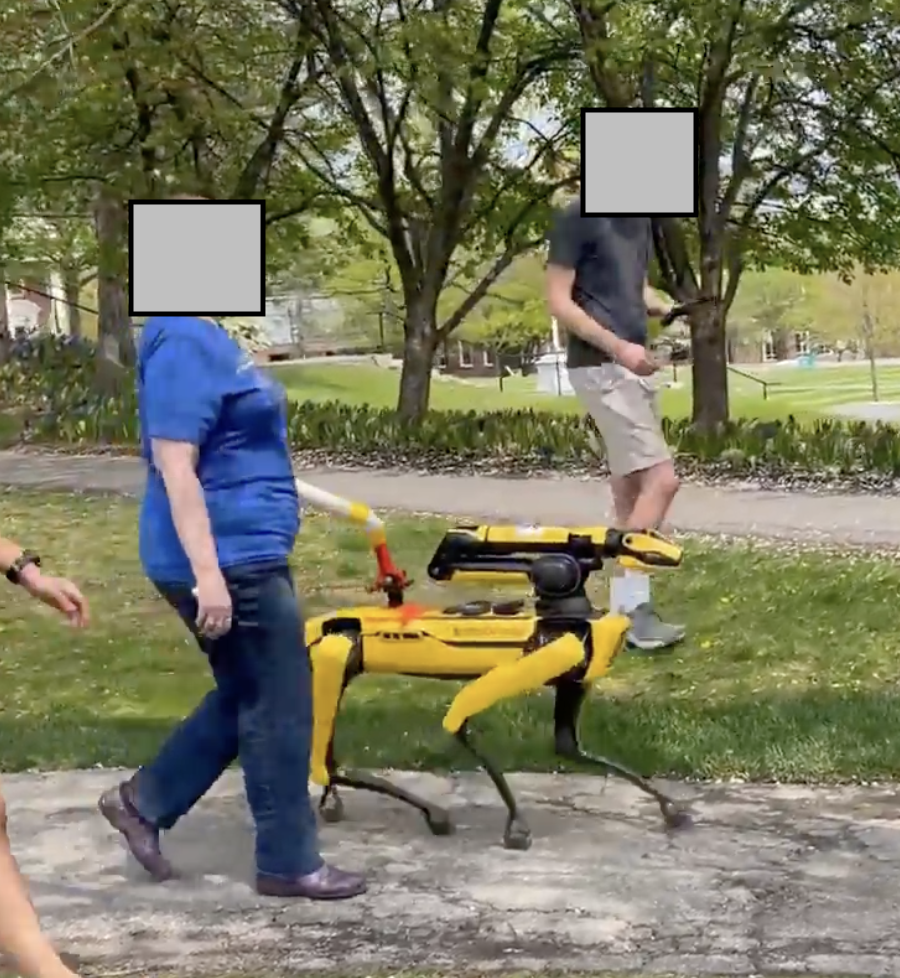}}
    \caption{Outdoor Field Test}
    \label{fig:OutdoorTest}
\end{figure}

Our BLV co-designer helped design and test the prototype study protocol and surveys that were given to the cited participants. Additionally, our co-designer participated in an autonomous walk test around a large square sidewalk loop. (Figures \ref{fig:OutdoorTest}). 
She reported that the handle was comfortable, and simulated the same walking, turning, and stair movements as those with her guide dog. The preliminary findings of her user test suggest that industrial agile robots are able to replicate many of the basic tasks that guide dogs perform. 

\begin{table}[ht]
\centering
\caption{Comparison of Guide Dog and Spot Explorer Navigation Abilities}
\begin{tabular}{|l|c|c|}
\hline
\textbf{Basic Guide Behaviors/Tasks} & \textbf{Guide Dog} & \textbf{Spot Explorer} \\
\hline
Lead handler in a straight line & X & X \\
\hline
Avoid obstacles and people & X & X \\
\hline
Maintain a consistent pace & X & X \\
\hline
Smooth and sharp turns & X & X \\
\hline
Walking backward (if needed) & X & X \\
\hline
Cross street at crosswalk & X & possible \\
\hline
\textbf{Object Recognition Tasks} & & \\
\hline
Crosswalks & X & possible \\
\hline
Overhead/Ground obstacles & X & possible \\
\hline
Changes in elevation & X & X \\
\hline
Object retrieval & X & X \\
\hline
\textbf{Verbal Commands} & & \\
\hline
Halt/Stop & X & X \\
\hline
Forward/Backward & X & X \\
\hline
Right/Left & X & X \\
\hline
Find (retrieval) & X & X \\
\hline
Sit/Down & X & X \\
\hline
\textbf{Other Training} & & \\
\hline
Intelligent Disobedience & X & possible \\
\hline
Ignoring Distractions & X & possible \\
\hline
\end{tabular}
\label{tab:RobotVDog}
\end{table}

\subsection{Navigation Task Results with Handle}
All of the participants were able to successfully complete: 1) the practice tasks with assistance, 2) the experimental handle navigation tasks unassisted without difficulty allowing the robot to guide them through a predetermined multilevel route, 3) the voice-based interface command tasks. The survey data assessed the levels of difficulty that participants reported while performing the protocol tasks.

\subsection{Handle Prototype Survey}
\begin{figure}
    \centering 
    \includegraphics[width=\linewidth]{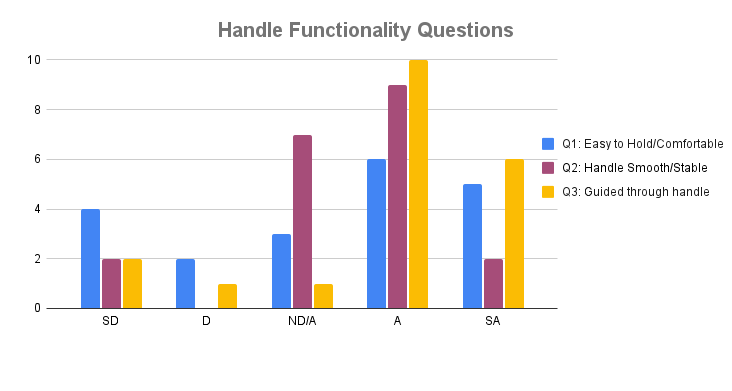} 
    \vspace{-1cm}
    \caption{Handle Survey Responses}
    \label{fig:HandleSurvey}

\end{figure}

Questions 1-3 asked participants about their experience using the prototype handle with the robot moving through the navigation tasks. Overall, the participants reported that they found the handle easy to hold and comfortable, and they could feel the robot guiding them through the handle (Figure \ref{fig:HandleSurvey}). There was less consistent agreement about the smoothness and the stability of the handle based on the large number of neutral responses to Question 2.

There was an additional open-response question asking for feedback about the functionality of the handle and how it could be improved. Most of the quotes mentioned the need for a better grip on the handle, which may not be positioned correctly to go down the stairs smoothly.

\begin{quote}

"For a makeshift handle, it was good. Maybe make it more ergonomic for the hand and slightly longer so it’s closer to the body."
\vspace{3mm}

"I feel like when walking down the stairs, the pull was a bit too abrupt and strong."

\vspace{3mm}
"It is a little awkward to hold and there are like dead spots where you don’t get any feedback from it at all so you can’t really tell where it’s going." 
\end{quote}

\begin{figure}
    \centering 
    \includegraphics[width=\linewidth]{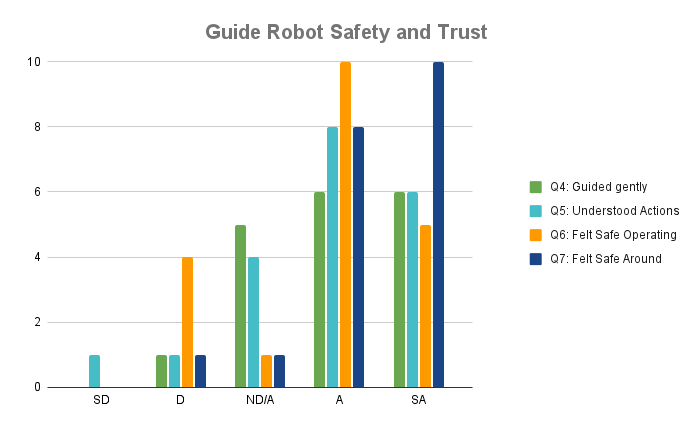} 
    \caption{Robot Safety Survey Responses}
    \label{fig:SafetySurvey}
\end{figure}

Questions 4-7 asked about the participant’s perceptions of safety and trust while being guided by the robot (Figure \ref{fig:SafetySurvey}). Overall, participants reported a high level of trust in the robot’s ability to guide them and that they felt safe both just being around the robot and operating the robot during the navigation tasks. However, there were several participants who reported they had safety concerns in the open response comments.

\begin{quote}
“The robot is just a bit clumsy overall. I don't think Spot was meant for guiding people.”

\vspace{3mm}
“Going down the stairs is difficult because when it does the bend to signal stairs you can't tell if it's going to go up or down, especially since you are disoriented because you have to rotate (the handle) around.”
\end{quote}

\subsection{ Voice-based Interface Prototype}
The results of the voice interface survey suggest that participants reported they could effectively pass simple movement commands (forward, backward, right, left) to the robot through the voice interface app. 

\begin{figure}[!h]
    \centering 
    \includegraphics[width=\linewidth]{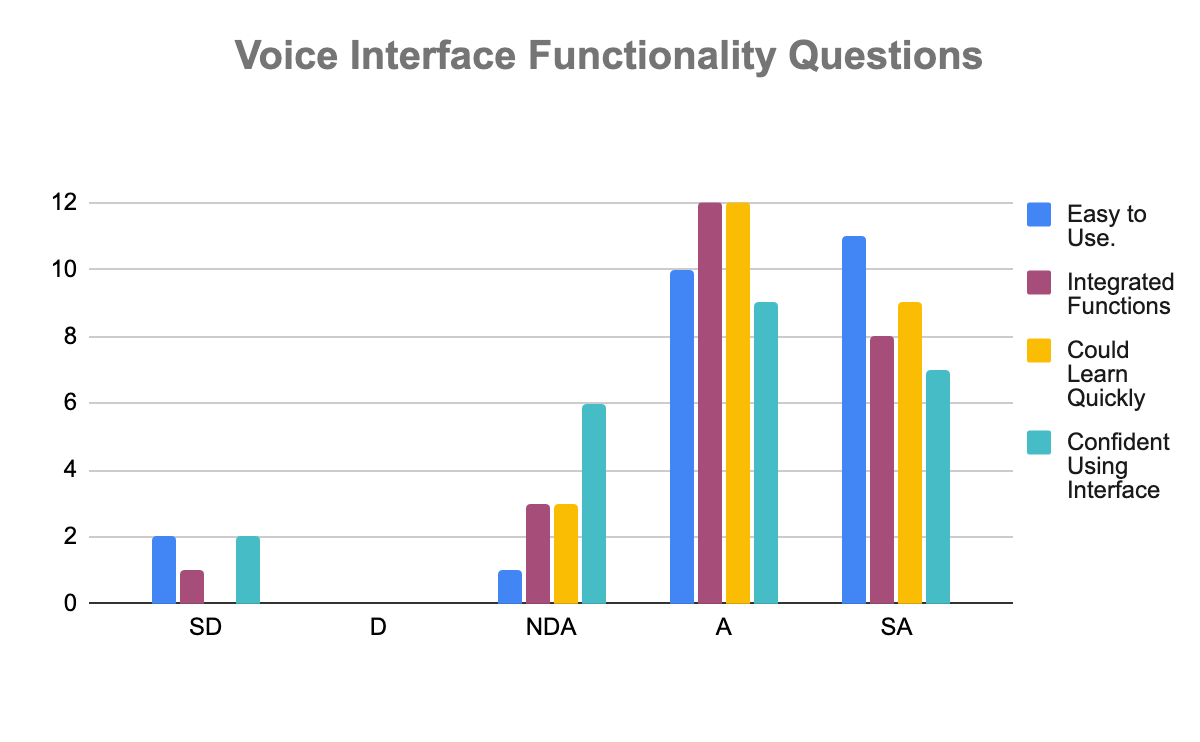} 
    \vspace{-1cm}
    \caption{Voice Interface Functionality }
    \label{fig:VoiceSurvey}
\end{figure}

\section{Discussion}
This study investigated the potential application of a commercially available agile industrial robot in a human-robot pair to perform a set of simple assistive navigation tasks that would be typical of a handler-guide dog pair. The next section will discuss the functioning of the two prototypes in the navigation tasks and preliminary results of the sighted participant study addressing the research questions motivating this study.

\textit{What are the necessary features for a multisensory interface to facilitate critical human-robot interactions for blind navigation in a multilevel indoor environment?}

We found that all of the participants in our human-robot pairings could perform similar simple guided navigation tasks as we observed in human-guide dog teams. The prototype handle proved to be an effective way for participants to receive information about the robot’s body motions to allow them to understand and feel comfortable with being led through the study route and tasks.  Participants indicated this was also true as the robot guided them up and down the short flight of stairs even though the robot needed to pivot around in order to go down backward correctly. The participants indicated the handle rotation allowed them to move around the robot as it pivoted on the landing. This preliminary study did not utilize the wide range of built-in features or navigation functionality available in this robot, nor did we implement the extensive object detection models that would allow for complex autonomous routes to be tested. Instead, we focused our attention on developing two early prototypes for the human-robot pair to communicate with one another through nonvisual channels (handle and the app).

Based on the feedback from our co-designer and the results of prototype usability studies, we argue there several benefits to the futures development of industrial quadruped robots for human-robot navigation applications. First, these robots (after further development) may alleviate the long waiting periods for BLV handlers while they are waiting for an appropirate guide dog match and allow the handler to learn new routes with more spatial information provided by the robot's internal navigation systems before learning new routes with their guide dog. This pre-journey training done with a human-robot pair could improve handler confidence when working with a new guide dog that is not currently possible during the critical transition periods between guide dog. Furthermore, the capacity for standardized, learned behaviors and consistency of mobility tasks can be adjusted and re-calibrated as needed. This type of industrial robot would be able to receive regular software updates, extending the potential working lifespan to meet the changing needs of the handler over the natural aging process. The handle prototype was effective for forward motion navigation tasks and did provide functional communication about the robot's guiding movements for the handler to anticipate and react. The early prototype for the voice-based interface also provided a way for the handler to communicate basic commands to the robot. The robot reacted to the commands with the appropriate actions for most of the trials. The ability of the human-robot pair to communicate through a voice-based interface has the potential to provide a greater level of spatial information for the handler than is currently available from a guide dog. The industrial robot used in this study has the capacity to receive, store, and communicate spatial information in the same manner as current accessible navigation applications. After further development, the voice-based app could communicate critical landmarks, relevant obstacles, and available route affordances to the handler along unknown routes. 

As with any study, our results also identified some challenges to implementing the necessary behaviors in human-robot guide interactions. Any size quadrupedal robot, small or large, is battery-operated, and the batteries need to be charged and replaced. Since we are using a larger and more sophisticated robot in this study, the battery life was a significant limitation and required planning to not run out of power during the user studies. The Spot Explorer model is currently limited to about 90 minutes of operating time before needing a battery replacement. This would make long trips or unexpected delays difficult for the handler in the real world. The batteries are about 15 lbs, making them too large to carry on longer walks. Furthermore, if this robot breaks or malfunctions, the repair would require a specialized engineer. While this might be a similar situation to guide dogs needing to visit the veterinarian when they are sick, the logistics of arranging for a 'house call' from a robotics specialist is currently impractical and would result in a long waiting time for a handler to get back to a daily routine. Finally, the Spot Explorer model has built-in safety features such as obstacle avoidance that currently interfere with effective human-robot navigation in environments where there may be crowds or other people in close proximity to the handler-robot pair. The Spot Explorer model is programmed to avoid objects if they get within approximately 8 inches to it's many cameras. While it can be customized to reduce that distance to 10cm, guide dogs are trained to 'part' crowds of people for their handler by pushing through and against people if given the command to do so. The current safety features interfere with the robot's ability to adjust to sudden changes in the environment, and it will stop to avoid injuries to others instead of doing the task that it is given by its handler.

\textit{What human-dog pair navigation behaviors and tasks can be reproduced with an industrial quadruped robot?}

For blind navigation in a multilevel indoor environment, critical human-robot communication interactions are vital in ensuring safe and effective movement for individuals with vision loss. The minimally necessary interactions between a human-robot pair were observed to be understanding of each other's movement intentions, the ability to avoid danger and obstacles, and making decisive decisions for the user. 
The Spot Explorer model is able to perform such tasks to varying degrees. With respect to communication between the user and Spot, there is good one-way communication where Spot is able to guide the user effectively through the handle. The participants of the user studies were able to understand Spot’s actions. However, beyond the movements from the handle, there are minimal forms of communication from Spot to the handler.  Furthermore, Spot cannot take commands from the handler, and all movements must be controlled manually or have instructions given beforehand. 

Guide dogs must make active decisions for the handler in the event that the handler’s actions may result in harming them; this is known as intelligent disobedience. For example, if the handler wants to cross a street, but the guide dog senses that there is a car coming, it will decide not to cross the street. The Spot Explorer model (or any other robot used in this type of research) does not currently have the ability to detect and communicate unknown dangers or react beyond its built-in programming in such a way. Therefore, the research and development of an agile industrial robot for navigation assistance still has a long way to go before it is ready for deployment in a real-world setting. However, based on these early explorations of commercially available technology, the results of this study suggest this is a viable research pathway that could present valuable co-design research opportunities for future versions of these robots. Perhaps with the cooperation and collaboration of industry partners, this may be a potential alternative to human-dog teams when situations or environments do not permit the use of trained guide dogs for safety or health reasons.

Although the Spot Explorer shows potential to be introduced as a robotic guide dog, future studies must address some limitations to expand the range of possible navigation behaviors. For example, there is only a one-way voice communication channel between the Spot Explorer and the handler. As found in the survey, the user is able to understand Spot’s actions through the handle, however, the robot does not always understand the handler’s intentions or is able to take real-time directions beyond the few commands we tested in this pilot study. In a typical handler-guide dog partnership, there is constant communication where the guide dog processes directions from the handler such as finding railings or turning in a certain direction. In response to the handler, the guide dog will follow directions, or if they notice obstacles, will lead the user around them. While the Spot robot can communicate with the user via the handle, it cannot provide more detailed spatial information for the handler to know why it might be avoiding obstacles. 

Due to the various cameras needed for navigation and sensing of the environment, there are limited positions where a handler can stand before they interfere with the cameras. Although we worked with our co-designer to adjust the handle to provide a comfortable position that was within the typical range for handler-dog pairs, if a handler were to accidentally block a camera on the side of Spot, it could begin to move erratically to avoid the handler. There were minimal problems while the handler performed simple tasks such as moving in a straight line; complex tasks such as pivoting on the stairs or turning often caused the user to block the cameras. In our user studies, the sighted participants were able to correct their position as they moved along, and our co-designer could easily adjust her position but this could prove to be a difficult or unsafe issue for other BLV handlers.

\section{Limitations}
While this preliminary study provided valuable information, it had several limitations. First, we only had 20 sighted participants recruited for the study. We focused on sighted participants for the interface prototype testing and for safety testing because we were not confident the robot would reliably be able to lead or swivel around easily going up or downstairs stairs with someone holding the handle. The small number of participants makes it difficult to generalize conclusions. Furthermore, there may be some inherent bias for the questions regarding trust in robot safety as the participants in the study were volunteers, and some had seen the robot previously being used in research on campus. Finally, the majority of our participants were sighted users who were not familiar with standard guide dog-handler communication techniques. Therefore, they were unable to provide a full range of feedback on the effectiveness of the communication through the handle as a part of the human-robot pair in the same way as our BLV co-designer. In addition, as they were sighted, they could maneuver themselves around obstacles or step away from the robot if the robot was trying to avoid obstacles, thus increasing their sense of control and safety in a set of simple navigation tasks. This problem is most prominent when Spot transitions from going up to down the stairs. Spot’s inability to go down the stairs forward requires Spot to pivot and for the user to switch hands or pivot along with Spot. There were several instances during the user studies where the user would block Spot’s camera once while transitioning and Spot would try to move away from the user. As the user is expected to move with Spot, Spot is unable to move away from the user to reset its position. There are also several improvements that could be made to adjust Spot’s pacing and acceleration when going from stop to start. The robot's pace would need to change depending on the scenario. Spot’s acceleration from stopping to start was often sudden. This is due to the power of the motors and the acceleration required to start and stop this large and powerful robot. This is something that can be adjusted in future studies.

\section{Conclusions and Future Work}
This research focuses on exploring an industrial agile quadruped robot as a potential alternative navigation assistant for BLV individuals. We investigated the larger, industrial Boston Dynamics Spot Explorer model as a viable option as a navigation assistant and how it might differ from similar research using smaller agile robots based on its ability to approximate typical guide dog behaviors and navigation tasks using two nonvisual interface prototypes for communication between the handler-robot pair. First, we worked with our co-designer to identify an industrial quadruped robot's potential benefits and limitations, and human-robot pair behaviors necessary for simple blind navigation tasks. We found that the Spot Explorer model could reproduce simple navigation tasks similar to an experienced guide dog, and the interface prototypes provided a basic level of communication and interaction between the handler-robot pair. However, we also identified several significant drawbacks based on the commercially available robot's current design and existing programming. The results of the prototype user studies suggest that there is evidence of adequate human-robot interactions and that users were able to understand the robot's actions through the handle. However, there is potential for significant improvement in the handle design and the addition of features such as haptics and switches to provide an immediate stop feature. Future design work and modifications will include changing the shape of the upper part of the handle, modifying the attachment structure of the handle to the robot, integrating object detection and retrieval, and an improved two-way voice-based interface to provide increased spatial information available to the handler through the robot's cameras and sensors.

\section*{Acknowledgement}
We would like to thank our labmates at the INSITE Lab at Colby College, the study participants for their time and feedback, and Toto, our expert guide dog. We would also like to thank Ed Colp at Boston Dynamics for supporting this project, without whom this research would not have been possible.


\nocite{*}  
\bibliographystyle{IEEEtran}
\bibliography{IEEEfull,refs}

\end{document}